\documentclass{article}

\usepackage{arxiv}

\usepackage[utf8]{inputenc} % allow utf-8 input
\usepackage[T1]{fontenc}    % use 8-bit T1 fonts
\usepackage{hyperref}       % hyperlinks
\usepackage{url}            % simple URL typesetting
\usepackage{booktabs}       % professional-quality tables
\usepackage{amsfonts}       % blackboard math symbols
\usepackage{nicefrac}       % compact symbols for 1/2, etc.
\usepackage{microtype}      % microtypography
\usepackage{lipsum}		% Can be removed after putting your text content
\usepackage{graphicx}
\usepackage{natbib}
\usepackage{doi}
\usepackage{amsmath}
\usepackage{multirow}
\usepackage{wrapfig}

\title{Planning or Learning: Reliability and Cost in Multi-Asset Maintenance}

\author{ Xian Yeow Lee \quad Chandrasekar Venkatraman \quad Ahmed Farahat \\ Industrial AI Lab, Hitachi America Ltd. \\ Santa Clara, U.S.A. \\ \texttt{\{xian.lee, chandrasekar.venkatraman, ahmed.farahat\}@hal.hitachi.com} }

\date{}

\renewcommand{\headeright}{}
\renewcommand{\undertitle}{}

\hypersetup{
    pdftitle={Planning or Learning: Reliability and Cost in Multi-Asset Maintenance},
    pdfsubject={Multi-Asset Maintenance, Predictive Maintenance, Reinforcement Learning, Planning},
    pdfauthor={Xian Yeow Lee, Chandrasekar Venkatraman, Ahmed Farahat},
    pdfkeywords={Multi-Asset Maintenance, Predictive Maintenance, Reinforcement Learning, Planning},
}

\begin{document}
\maketitle

% Abstract
\begin{abstract}
Industrial maintenance systems increasingly involve multiple interacting assets and shared resources, making it challenging to balance reliability and operational cost using a single decision framework. While recent work has focused on reinforcement learning (RL) for maintenance scheduling, direct comparisons with planning approaches under identical settings remain limited. In this work, we empirically compare planning and RL for multi-asset bearing maintenance using run-to-failure data. We examine how these methods behave when balancing preventive maintenance against tolerable failures across a range of failure penalty scenarios. We observed a consistent behavioral difference driven by objective formulation. Planning enforces reliability as a hard constraint and produces zero-failure policies whose total cost is largely insensitive to the magnitude of failure penalties. RL agents optimize expected cost and often trade off preventive maintenance against occasional failures as penalties vary, resulting in lower costs under low-penalty regimes but persistent non-zero failures even when penalties are high. We also investigate lightweight constraint mechanisms, including reward shaping and action masking, to encourage RL's reliability. From a practical perspective, planning may be more suitable when strict reliability is required and deployment horizons are short, whereas RL may provide cost-efficient policies when limited failures are acceptable and long-run operational efficiency is prioritized. Overall, this study clarifies the trade-offs between reliability and cost in multi-asset maintenance, highlights the challenges of enforcing zero-failure behavior in RL, and suggests that planning and RL are complementary approaches whose applicability depends on operational objectives. Beyond these findings, the controlled benchmark protocol itself that unifies environment, cost model, and evaluation across paradigms, offers a reusable template for comparing decision-making approaches in other maintenance settings.
\end{abstract}

\section{Introduction}
Industrial equipment increasingly operates in fleets where multiple assets degrade simultaneously and share maintenance resources such as technicians, tools, and scheduled downtime windows \cite{o2025optimizing}. For example, in wind farms, taking one turbine offline for maintenance can require reallocating crews and coordinating downtime across nearby units, directly affecting the availability and risk exposure of the remaining turbines. Similar coordination challenges arise in rotating machinery, where multiple bearings degrade along independent run-to-failure trajectories but must be serviced by a shared maintenance team. In such settings, decisions are inherently coupled, and maintenance planning becomes a fleet-level scheduling problem rather than a collection of independent asset-level decisions.

Two broad classes of approaches have been applied to this class of problem. Planning methods formulate the scheduling task explicitly, searching over a joint fleet state with a defined objective and, optionally with hard constraints on failures or downtime. Reinforcement learning (RL) methods instead learn a scheduling policy through interaction, encoding the problem as a Markov decision process and optimizing a discounted sum of rewards. These methods have recently gained popularity in part because they can naturally handle uncertainty in degradation dynamics and decision outcomes without requiring an explicit model of all system transitions \cite{RLReview1}. Both approaches have demonstrated value in maintenance and related sequential decision-making contexts. However, controlled comparisons between planning and RL under a shared environment and cost structure remain rare, even in adjacent scheduling domains; with one exception being a study of unit commitment for power plant scheduling, where mixed-integer programming was shown to consistently produce lower-cost, more reliable schedules than RL under matched conditions \cite{omalley2023rlmip}.

While both approaches have similarities, the comparison is nontrivial because they do not solve the same problem. A planning method with a hard zero-failure constraint will always produce a zero-failure schedule if one is feasible, making the magnitude of the failure penalty irrelevant to its decisions. RL paradigms, by contrast, commonly treat failure as a soft cost that it may accept if the discounted benefit of prevention is smaller than the cost of maintenance. This difference in objective formulation can produce systematically different policies even when both methods have access to the same state information.

This paper presents a systematic empirical comparison of planning and RL for an instance of multi-asset maintenance scheduling using a run-to-failure bearing dataset. In this paper, we make the following contributions: 
\begin{itemize} 
\item A controlled empirical benchmark that places planning and RL under a unified environment, cost structure, and evaluation protocol for multi-asset maintenance scheduling, providing both a direct comparison between the two paradigms and a reusable template for evaluating decision-making approaches in other maintenance and PHM settings.
\item A characterization of a formulation-driven behavioral gap: planning enforces failure avoidance as a hard constraint and is invariant to the failure penalty magnitude, while RL optimizes a discounted soft cost and exhibits a structural bias toward zero-maintenance when asset lifetimes exceed the effective discount horizon. 
\item An assessment of lightweight reliability mechanisms showing that reward shaping for RL has negligible effect on failure behavior, action masking reduces failures more substantially but at significant cost overhead, but neither closes the reliability gap to planning. 
\end{itemize}

\section{Related Work}

\subsection{Planning for Maintenance Scheduling}

Condition-based maintenance (CBM) and predictive maintenance (PdM) use health state estimates to schedule interventions before failures occur \cite{sakib2018challenges}. For multi-asset systems, scheduling involves allocating shared maintenance resources across assets with heterogeneous degradation states, which introduces dependencies that single-asset policies cannot account for \cite{petchrompo2019review}. Rule-based policies such as fixed-interval replacement and state-threshold triggering are common in practice due to their simplicity, but they do not adapt to the joint fleet state and may over- or under-maintain individual assets.

Model predictive control (MPC)-type rolling-horizon planning approaches have also been applied to maintenance scheduling \cite{zhang2025integrated}. These methods formulate the scheduling task as a sequential decision problem over an explicit state space, enabling joint optimization across all assets. A common execution strategy is receding-horizon control, where only the first action of the planned sequence is executed and the plan is recomputed at each step from the updated state. Nonetheless, the computational cost of online re-planning is a practical limitation at scale.

\subsection{RL for Maintenance Scheduling}

RL has received increasing attention for maintenance scheduling because it can learn cost-minimizing policies from interaction without requiring an explicit system model \cite{RLPdm1, RLPdm2, RLReview1}. Extensions to multi-asset settings with shared crew or budget constraints are increasingly being investigated, though systematic comparisons with planning baselines under the same environment remain scarce \cite{zhang2022deep}. A recurring challenge in applying RL to maintenance is reward sparsity: the consequence of deferring maintenance (a failure) may not materialize for hundreds or thousands of steps, making credit assignment difficult \cite{pignatellisurvey}. This challenge is compounded in multi-asset settings where the crew constraint creates additional dependencies between action consequences.

Several approaches address the problem of enforcing reliability constraints in learned policies. Reward shaping amplifies penalty terms during training to steer the policy toward constraint satisfaction, without modifying the policy architecture or evaluation protocol \cite{hu2020learning}. Action masking prevents the selection of actions violating a hand-crafted safety rule, reducing the feasible action space at training and deployment time \cite{hou2023exploring}. Constrained Markov decision processes (CMDPs) provide a principled framework for enforcing inequality constraints on expected cumulative costs, with Lagrangian relaxation as a standard solution method \cite{altman2021constrained}, with recent work extending these guarantees to safe exploration during training and to stochastic stopping-time settings \cite{wachisui2020safe, mazumdar2024safe}, and to constrained POMDP formulations for inspection and maintenance planning specifically \cite{andriotis2021deep}. Shielding approaches impose a formal safety monitor between the agent and environment, overriding unsafe actions at runtime based on a specification \cite{alshiekh2018safe}. In this work we evaluate reward shaping and action masking as lightweight approaches and leave CMDP and shielding as future studies.

Beyond CMDPs and shielding, other formulations aim to reduce the gap between RL's objective and reliability-oriented planning. Average-reward RL optimizes long-run average cost instead of geometrically discounted return, which avoids the discount-horizon mismatch discussed in Section~\ref{subsec:dqn_analysis}
; similar horizon-related biases in discounted RL have been documented in other long-horizon sequential decision-making domains \cite{wireless2025average}. Risk-sensitive RL instead incorporates an explicit risk measure, such as conditional value-at-risk or exponential utility, into the training objective or as a constraint, offering another route to bias policies away from rare but costly failure events \cite{prashanth2022risk}. We do not evaluate these formulations here but view them, alongside CMDPs, as promising directions for narrowing the reliability gap observed between planning and RL in this study.

\begin{figure*}[h!]
    \centering
    \includegraphics[width=\linewidth]{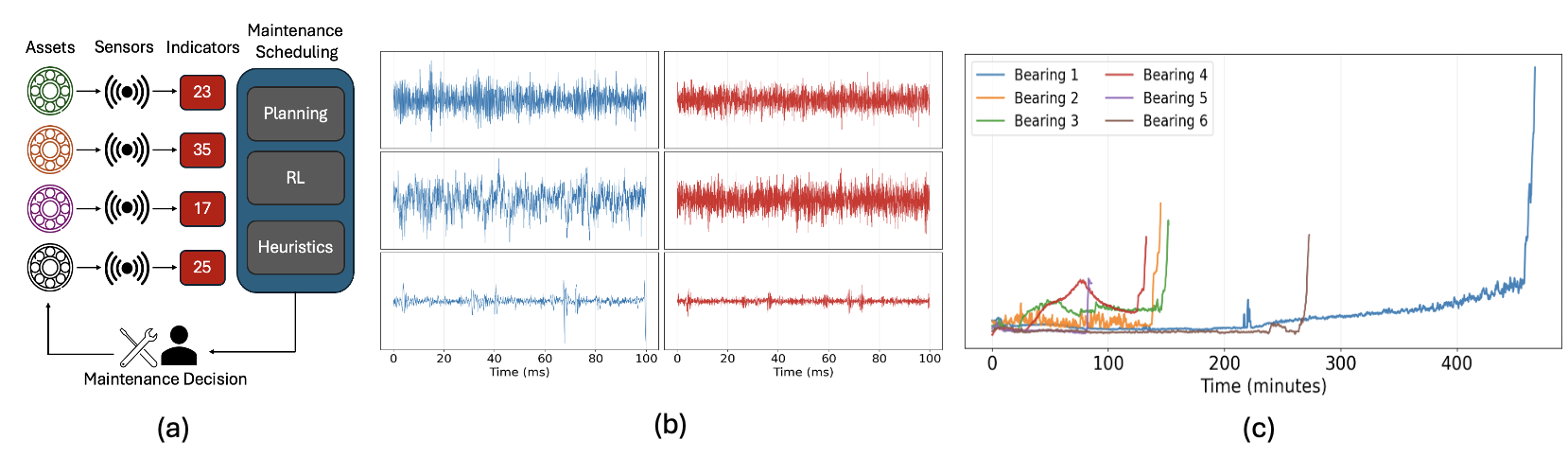}
    \caption{(a) Pipeline from raw vibration signals to health indicators to maintenance decisions. (b) Raw horizontal (blue) and vertical (red) accelerometer signals for a bearing at early stages (row 1), mid-life (row 2), and near-failure stages (row 3). (c) Composite RMS health indicator trajectories for all six bearings, each following a distinct degradation profile before failure.}
    \label{fig:pipeline}
\end{figure*}

\section{Problem Formulation}

\subsection{Multi-Bearing System}\label{subsec:dynamics}

We consider a fleet of $N$ bearings operating in parallel, each following an independent degradation trajectory toward failure. To concretize this, we illustrate the typical pipeline in Fig.~\ref{fig:pipeline}(a), where raw sensor signals, shown in Fig.~\ref{fig:pipeline}(b) are typically processed into estimated RULs or discrete health indicators. Based on the observed readings, a maintenance schedule is generated either manually, via a fixed rule, or through a planning or RL algorithm in this work. A single maintenance crew is shared across all bearings, so at most one bearing can be serviced per time step. When a bearing receives maintenance it is restored either to new-equivalent condition or an estimated new life. All other bearings continue to degrade normally during a maintenance step. 

Let $s_t = (h_t^1, \dots, h_t^N)$ denote the fleet state at time $t$, where $h_t^i$ is the health or remaining useful life of bearing $i$. Let $a_t \in \{0, 1, \dots, N\}$ denote the action, where $a_t = 0$ corresponds to no maintenance and $a_t = i$ indicates that bearing $i$ is serviced. The system evolves according to
\begin{equation}
h_{t+1}^i =
\begin{cases}
g(h_t^i), & \text{if } a_t \neq i \\
\tilde{h}_0^i, & \text{if } a_t = i
\end{cases}
\end{equation}
where $g(\cdot)$ denotes the degradation dynamics and $\tilde{h}_0^i$ is the post-maintenance state, which may correspond to a new or estimated lifetime. The goal of the multi-asset maintenance problem is to come up with an optimal plan to maintain the bearings, while incurring minimum cost and failures, subject to the constraint of crew availability.

\subsection{Mapping to Discrete Health Indicator}\label{subsec:hi}

A common practice in maintenance scheduling is to map raw sensor data to discrete health indicators to simplify the methods. In this work, we explore two variants, where the planning and RL algorithm either observes the estimated continuous RUL or a discrete state of the system. 

To map raw sensor data to discrete states, we do the following: For each bearing, we compute a composite health indicator from horizontal and vertical vibration accelerometer signals. At each 10-second recording window, the root mean square (RMS) amplitude is computed over the 2560 available samples for each channel. The composite health indicator is then defined as:

\begin{equation}
    \text{HI}(t) = \sqrt{\text{RMS}_h(t)^2 + \text{RMS}_v(t)^2}
    \label{eq:hi}
\end{equation}

where $\text{RMS}_h$ and $\text{RMS}_v$ denote the horizontal and vertical channel amplitudes, respectively. A moving average is applied to suppress measurement noise. The health state of each bearing is then discretized into one of four categories: \textit{healthy, degrading, critical}, or \textit{failed}, using ground truth remaining useful life labels: a bearing is \textit{healthy} if it retains more than 75\% of its initial lifetime, \textit{degrading} between 25\% and 75\%, \textit{critical} below 25\%, and \textit{failed} at zero remaining life.

\subsection{Cost Model}
\label{subsec:costmodel}

We define the operational cost of an episode with $T$ steps as:

\begin{equation}
    C = \sum_{t=1}^{T} \Bigl[ c_\text{op} \cdot N_\text{act}(t) + c_\text{m} \cdot \mathbf{1}[a_t \neq \text{noop}] + c_\text{f} \cdot N_\text{fail}(t) \Bigr]
    \label{eq:cost}
\end{equation}

where $c_\text{op} = 1.0$ is the per-step operating cost per active bearing, $c_\text{m} = 25.0$ is the fixed cost of a maintenance action, $c_\text{f}$ is the failure penalty (varied across experiments), $N_\text{act}(t)$ is the count of non-failed bearings that are actively operating at step $t$, excluding the bearing, if any, currently under maintenance: a bearing being serviced is treated as down for that step and does not accrue operating cost, $N_\text{fail}(t)$ is the count of bearings that newly fail at step $t$, and $a_t$ is the action taken at step $t$. The failure cost $c_\text{f}$ is the primary free parameter representing an operator's risk tolerance: a low value permits occasional failures in exchange for lower maintenance expenditure, while a high value reflects a safety-critical regime where failures are strongly penalized. All other cost parameters are fixed across all experiments. This cost parameter is used as the objective in both planning and RL experiments.

\subsection{Action Space and Scheduling Constraint}

At each step, the planning / RL algorithm selects an action from $\mathcal{A} = \{ \text{no-op}, \text{maintain } B_1, \ldots, \text{maintain } B_N \}$, where $N = $ number of bearings. The no-op action advances the system one step without intervention. A maintenance action assigns the crew to service the selected bearing for one step while all other bearings continue to operate. The crew constraint means that only one bearing can be maintained per step, and a bearing whose maintenance is deferred while in the failed state incurs no operating cost.

\subsection{Post-Maintenance Dynamics}\label{subsec:post-dynamics}

We evaluate two models of post-maintenance bearing lifetime that are observed by the planning and RL algorithms. Under oracle dynamics, servicing bearing $B_i$ resets its trajectory to step zero of its original recorded run, so the bearing follows the same degradation sequence as its initial life. This creates a deterministic setting in which future degradation is perfectly repeatable, eliminating uncertainty in post-maintenance behavior and providing a controlled test of scheduling performance. Under estimated dynamics, the post-maintenance lifetime is instead sampled from a log-normal distribution fitted to the observed lifetimes of all six bearings. This introduces stochasticity and model mismatch between expected and realized bearing lifetime, more closely reflecting practical deployment conditions where the future behavior of a repaired component is uncertain.

\section{Methods}

\subsection{Planning: Dijkstra Search}

In this work, we use Dijkstra's algorithm \cite{dijkstra1959note} as a representative planning approach. Maintenance scheduling is formulated as a finite-horizon shortest-path problem over a discrete joint state space. Each state is a tuple $s = (t,\, \hat{r}_1, \ldots, \hat{r}_N,\, f_1, \ldots, f_N)$, where $t$ is the current step, $\hat{r}_i$ is the remaining life of bearing $i$ (capped at $H+1$ to maintain tractability), and $f_i \in \{0, 1\}$ indicates whether bearing $i$ has failed.

A hard zero-failure constraint is imposed by assigning infinite cost to any state in which a bearing failure occurs. The planner searches for the minimum-cost action sequence over the next $H$ steps that avoids all failures using uniform cost search. Only the first action is executed, then the plan is recomputed from the new state in a receding-horizon manner. Because failure is enforced as a hard constraint rather than a weighted objective, the planner's decisions are independent of the failure cost parameter $c_\text{f}$: any zero-failure schedule is preferred over any schedule with failures regardless of its magnitude. This property is central to the comparison in Section~\ref{subsec:compare}.

When a bearing is maintained, its remaining life resets to the capped value and it does not age during that step, following the same downtime convention introduced in Section~\ref{subsec:costmodel}. As the search expands, only the lowest-cost path found to each state is retained, and any remaining ties are resolved through a fixed ordering over the state representation. We rely solely on the remaining-life cap described above to keep the search space tractable, without introducing further pruning across bearings. Expansion stops as soon as a horizon-length trace with zero failures is reached, which uniform-cost search guarantees to be optimal, and the planner reports infeasibility if no such trace exists within the horizon.

\subsection{Deep Reinforcement Learning}

We consider two widely used RL approaches: a value-based method, Deep Q-learning (DQN) \cite{dqn}, and a policy-based method, Proximal Policy Optimization (PPO) \cite{ppo}.

\textbf{Deep Q-Learning:} DQN approximates the action-value function $Q_\theta(s, a)$ using a neural network. During training, actions are selected using an $\epsilon$-greedy policy with decaying exploration, while evaluation uses greedy action selection. The network is trained by minimizing the squared temporal-difference error:

\begin{equation}
    \mathcal{L}(\theta) = \mathbb{E} \Bigl[ \bigl( r + \gamma \max_{a'} Q_{\bar{\theta}}(s', a') - Q_\theta(s, a) \bigr)^2 \Bigr]
    \label{eq:dqn}
\end{equation}

where $r$ is the immediate reward (negative of the step cost from Eq.~\ref{eq:cost}), $\gamma$ is the discount factor, $s'$ is the next state, and $Q_{\bar{\theta}}$ is a periodically updated target network used to stabilize training.

We define the observation space as a continuous representation, where each bearing contributes its normalized remaining useful life $r_i / R_i$ and a binary failure flag.

\textbf{Proximal Policy Optimization:} PPO directly optimizes a stochastic policy $\pi_\theta(a \mid s)$ using the clipped surrogate objective:

\begin{equation}
    \mathcal{L}^\text{CLIP}(\theta) = \mathbb{E}_t \Bigl[ \min \bigl( \rho_t \hat{A}_t,\; \text{clip}(\rho_t, 1 - \varepsilon, 1 + \varepsilon)\, \hat{A}_t \bigr) \Bigr]
    \label{eq:ppo}
\end{equation}

where $\rho_t = \pi_\theta(a_t \mid s_t) / \pi_{\theta_\text{old}}(a_t \mid s_t)$ is the probability ratio between successive policies and $\hat{A}_t$ is the generalized advantage estimate. An entropy bonus is included to encourage exploration. PPO operates on-policy, collecting trajectories under the current policy before performing updates. The same observation space used for DQN is also used for PPO.

\subsubsection{Constraint Mechanisms}

To encourage zero-failure behavior in RL agents, we investigate two commonly used approaches: reward shaping and action masking.

Reward shaping multiplies the failure cost term in the reward signal by a penalty amplifier $M$ during training only. Evaluation always uses the original cost parameters for a fair comparison. The goal is to produce a policy that avoids failures even at lower deployment penalties.

The action mask forbids the no-op action at any step where at least one non-failed bearing has a normalized remaining life below a threshold $\delta$:

\begin{equation}
    \text{mask noop if} \quad \exists\, i : (r_i / R_i < \delta) \wedge (f_i = 0)
    \label{eq:mask}
\end{equation}

When the mask is active, the agent must select a maintenance action, though the choice of which bearing to maintain remains unconstrained. We use $\delta = 0.25$ throughout, meaning the mask activates when any non-failed bearing has less than 25\% of its estimated remaining life. The mask is applied during both training and evaluation.

\subsection{Rule-Based Baselines}

We include a rule-based baseline as reference. The \textit{threshold} policy maintains the bearing that has most recently crossed a degradation threshold, scheduling maintenance as soon as the crew becomes available. When multiple bearings satisfy the threshold condition simultaneously, priority goes to the one with the smallest RUL; any residual tie is broken by fixed bearing order. The baseline is designed to avoid failures under oracle dynamics by scheduling maintenance conservatively. It requires no training and incurs negligible inference cost.

\section{Experimental Setup}

This section describes the experimental configuration used to evaluate all methods under a common environment and cost structure. Reproduction code is available at the linked repository. \footnote{\url{https://github.com/xylhal/PHM_PlanningVsRL}}

\subsection{Dataset and Environment}

We use a run-to-failure bearing dataset \cite{nectoux2012pronostia} as the basis for our experimental environment. This dataset is widely used for remaining useful life estimation and has supported both physics-based and data-driven prognostic methods \cite{dhungana2025bearing}.

We use the training portion of the data, which contains six run-to-failure bearing trajectories. The six bearings have lifetimes of 515, 797, 871, 911, 1637, and 2803 recording windows, where each window corresponds to a 10-second interval. This yields total lifetimes ranging from $\approx$ 86 to 467 minutes. A single episode simulates all six bearings operating in parallel using the transition dynamics of bearing degradation as described in Section~\ref{subsec:dynamics}. At each time step, at most one bearing can be serviced, reflecting a shared maintenance resource constraint. Episodes run for 8000 steps, exceeding the maximum observed lifetime to ensure that failures occur in the absence of maintenance.

We also conducted our experiments using the two types of post-maintenance dynamics as described in Section~\ref{subsec:post-dynamics} - the oracle dynamics and estimated dynamics - to evaluate performance under both idealized and uncertain conditions. 

The failure penalty is varied across $c_f$ =\{100, 500, 1000, 5000, 10000\}, spanning regimes where failure is relatively inexpensive to highly penalized compared to the maintenance cost of 25. All other cost parameters are held fixed, with $c_{\text{op}} = 1.0$ and $c_m = 25.0$.

\subsection{Planning and Heuristic Protocol}

For planning using Dijkstra, we use a finite horizon of $H = 20$, which balances computational cost with sufficient lookahead over future degradation. 

For heuristic, the \textit{threshold} policy applies maintenance when a bearing reaches the critical health state or worse. When multiple bearings satisfy the threshold condition, priority is given to the one closest to failure.

All methods operate under the constraint that at most one bearing can be serviced at each time step. Planning and heuristic baseline policies are deterministic, and therefore we only report a single run for each configuration.

\subsection{RL Training and Evaluation Protocol}

RL agents are trained for 1500 episodes (DQN) and 1000 episodes (PPO) across three independent random seeds ($\{0,1,2\}$, applied to Python, NumPy, and PyTorch RNGs, and also to the stochastic post-maintenance sampling under estimated dynamics). Unless otherwise stated, experiments use a discount factor of $\gamma = 0.99$. The default state representation is continuous, consisting of the normalized remaining life and failure indicator for each bearing, resulting in a state dimension of $2N$ for $N$ bearings. We also evaluate a discrete variant where each bearing is represented by a health-state index, giving a state dimension of $N$. The action space is discrete, with one action corresponding to no maintenance and one action per bearing. Actions attempting to service already failed bearings are masked during selection. 

Both DQN and PPO use a two-layer fully connected neural network with 64 hidden units per layer and ReLU activations, followed by a linear output layer. The reward is defined as the negative step cost, directly corresponding to Eq.~\ref{eq:cost}. DQN uses an experience replay buffer of size 50000, mini-batch size 64, learning rate $3 \times 10^{-4}$, and target network updates every 500 gradient steps. Gradient updates are performed every 16 environment steps, with exploration following an $\epsilon$-greedy schedule decaying from 1.0 to 0.05. PPO uses generalized advantage estimation ($\lambda = 0.95$), clipping parameter 0.2, entropy coefficient 0.01, mini-batch size 64, roll-out length 1024, and 4 gradient epochs per update. Both methods are optimized using Adam with learning rate $3 \times 10^{-4}$, and PPO gradients are clipped to a maximum norm of 0.5.

During training, checkpoints are saved every 200 episodes and evaluated over 5 episodes using greedy action selection. The checkpoint with the lowest mean evaluation cost is selected as the final model. Final performance is then evaluated over 10 episodes, reporting the mean and standard deviation across seeds.

For reward shaping experiments, the failure penalty during training is multiplied by $M = 10$ while evaluation uses the original reward function. We also evaluate an action-masking 
variant as described in Eq.~\ref{eq:mask}. All hyperparameters above were selected based on standard literature defaults (Adam, PPO's clipping and GAE parameters), while a small number of choices, such as learning rate and DQN's update frequency, were set manually for this environment.

\section{Results}
%HERE, need to caption
\begin{wrapfigure}{r}{0.5\textwidth}
    \centering
    \includegraphics[width=1\linewidth]{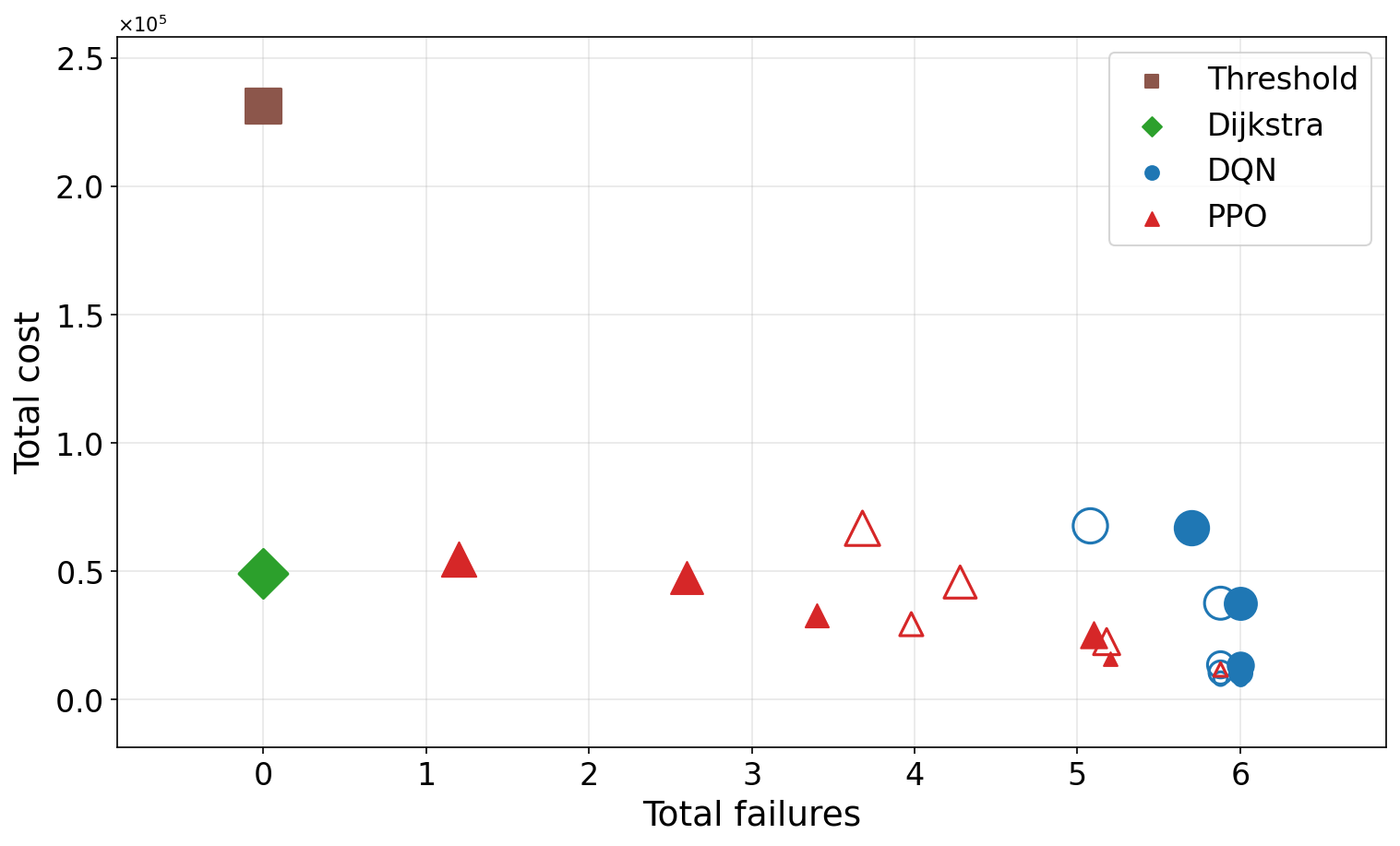}
    \caption{Pareto tradeoff between cost and failures across all methods and failure penalties $c_f$, where marker size $\propto c_f$ and filled = oracle dynamics, hollow = estimated dynamics. Planning and heuristic are anchored at zero failures; RL methods cluster in the high-failure, low-cost region regardless of $c_f$, reflecting the hard-constraint vs.\ soft-cost formulation gap.}
    \label{fig:pareto}
\end{wrapfigure}

\subsection{Main Comparison}\label{subsec:compare}

Table~\ref{tab:core} reports the overall comparison of heuristic, planning, and RL methods across different $c_f$ values and under both oracle and estimated post-maintenance dynamics. Across all experiments, heuristic (\textit{threshold}) and planning (Dijkstra) methods strictly avoid failures, while RL methods may trade off failures to reduce total cost. This difference follows directly from the formulation: heuristic and planning approaches enforce failure avoidance as a hard constraint, whereas RL optimizes a cost-based objective in which failures remain permissible. Hence, both heuristic and planning methods are effectively independent of $c_f$, while RL performance varies as a function of operating and failure costs. This distinction is important from a practitioner perspective, as it reflects whether failure avoidance is treated as a strict requirement or as part of a cost trade-off.

Comparing heuristic and planning approaches, Dijkstra is consistently more cost-effective than the \textit{threshold} policy. This highlights the benefit of explicitly searching over future system evolution, even within a constrained planning horizon, although practical deployment considerations such as computational cost are discussed later in Section~\ref{subsec:runtime}.

Comparing the RL methods, DQN achieves lower total cost than PPO across most configurations. However, this cost advantage is accompanied by a tendency toward higher failure counts, suggesting that DQN may converge to policies that under-utilize maintenance. While DQN exhibits near-zero variance across random seeds, this likely reflects convergence to a simple or degenerate policy rather than robustness. In contrast, PPO shows higher variability across seeds and generally fewer failures, though its sensitivity to the failure cost parameter is less consistent and varies across settings.

When comparing oracle and estimated dynamics, the relative ranking of methods remains largely unchanged, indicating that our observations are robust to uncertainty in post-maintenance behavior. Dijkstra maintains zero failures under both dynamics, with only a small reduction in cost under estimated dynamics due to occasional longer sampled lifetimes. RL methods exhibit similar qualitative behavior in both settings, suggesting that the observed trade-offs are not sensitive to the specific choice of post-maintenance model. 

Overall, the results show a clear trade-off between cost and reliability. Planning and heuristic methods can guarantee zero failures but incur higher cost, while RL methods can achieve substantially lower cost in low failure-penalty regimes by allowing failures. As the failure penalty increases, planning becomes increasingly favorable, eventually dominating RL in both cost and reliability. This pareto trade-off is illustrated in Fig.~\ref{fig:pareto}.

% -----------------------------------------------------------------------
% TABLE 1: CORE METHODS (Oracle + Estimated merged)
% -----------------------------------------------------------------------
\begin{table*}[h]
\centering
\caption{Main results for planning, heuristic, and vanilla RL methods across all failure penalty levels. Cost is the mean total episode cost; Failures is the mean number of bearing failures per episode. Standard deviation across three seeds is shown for RL methods. Dijkstra and \textit{Threshold} are deterministic (no standard deviation).} 
\label{tab:core}
\scriptsize
\setlength{\tabcolsep}{4pt}
\resizebox{\textwidth}{!}{%
\begin{tabular}{l | ll | ll | ll | ll | ll}
\toprule
\multirow{2}{*}{Method} &
  \multicolumn{2}{c|}{$c_\text{f} = 100$} &
  \multicolumn{2}{c|}{$c_\text{f} = 500$} &
  \multicolumn{2}{c|}{$c_\text{f} = 1000$} &
  \multicolumn{2}{c|}{$c_\text{f} = 5000$} &
  \multicolumn{2}{c}{$c_\text{f} = 10000$} \\
 & Cost & Failures & Cost & Failures & Cost & Failures & Cost & Failures & Cost & Failures \\
\midrule
\multicolumn{11}{l}{\textit{Oracle dynamics}} \\
\midrule
Threshold      & 231336          & 0.0       & 231336          & 0.0       & 231336          & 0.0       & 231336          & 0.0       & 231336          & 0.0       \\
Dijkstra       & 49176           & 0.0       & 49176           & 0.0       & 49176           & 0.0       & 49176           & 0.0       & 49176           & 0.0       \\
DQN       & 8134 $\pm$ 0    & 6.0 $\pm$ 0.0 & 10534 $\pm$ 0   & 6.0 $\pm$ 0.0 & 13534 $\pm$ 0   & 6.0 $\pm$ 0.0 & 37534 $\pm$ 0   & 6.0 $\pm$ 0.0 & 66856 $\pm$ 959  & 5.7 $\pm$ 0.5 \\
PPO       & 15918 $\pm$ 2561 & 5.2 $\pm$ 0.4 & 32973 $\pm$ 3751 & 3.4 $\pm$ 0.4 & 25330 $\pm$ 4721 & 5.1 $\pm$ 0.8 & 47427 $\pm$ 3293 & 2.6 $\pm$ 1.4 & 54780 $\pm$ 2542  & 1.2 $\pm$ 0.8 \\
\midrule
\multicolumn{11}{l}{\textit{Estimated dynamics}} \\
\midrule
Threshold      & 231336          & 0.0       & 231336          & 0.0       & 231336          & 0.0       & 231336          & 0.0       & 231336          & 0.0       \\
Dijkstra       & 48936           & 0.0       & 48936           & 0.0       & 48936           & 0.0       & 48936           & 0.0       & 48936           & 0.0       \\
DQN       & 8134 $\pm$ 0    & 6.0 $\pm$ 0.0 & 10534 $\pm$ 0   & 6.0 $\pm$ 0.0 & 13534 $\pm$ 0   & 6.0 $\pm$ 0.0 & 37534 $\pm$ 0   & 6.0 $\pm$ 0.0 & 67676 $\pm$ 201   & 5.2 $\pm$ 1.1 \\
PPO       & 11708 $\pm$ 1773 & 6.0 $\pm$ 0.0 & 29375 $\pm$ 2893 & 4.1 $\pm$ 0.7 & 22570 $\pm$ 2494 & 5.3 $\pm$ 0.3 & 45786 $\pm$ 4681 & 4.4 $\pm$ 0.8 & 66738 $\pm$ 1749 & 3.8 $\pm$ 0.9 \\
\bottomrule
\end{tabular}
}
\end{table*}

%Runtime
\subsection{Planning Cost and Runtime}\label{subsec:runtime}

\begin{table}[t]
\small
\centering
\caption{Training and inference runtimes for all methods.}
\label{tab:runtime}
\begin{tabular}{l r r}
\toprule
Method & Training & Inference (per episode) \\
\midrule
Threshold       & ---          & 28 s \\
Dijkstra        & ---         & $\sim$90 min \\
DQN             & $\sim$40 min & 9 s \\
PPO             & $\sim$68 min & 10 s \\
\bottomrule
\end{tabular}
\end{table}

Table~\ref{tab:runtime} reports training and inference runtimes for all methods. All experiments were conducted on a workstation with a 6-core Intel Xeon-class CPU and 32GB RAM, ensuring a fair comparison of planning and policy-based computation under CPU-bound conditions. Dijkstra incurs substantially higher inference cost, requiring $\approx$90 minutes per episode due to repeated re-planning over the full horizon. This cost is dominated by repeated search over the joint fleet state at every decision step. In contrast, heuristics and RL policies have negligible inference overhead, with heuristics requiring 28 seconds and with DQN and PPO requiring $\approx$ 9 and 10 seconds per episode, respectively.

The key distinction lies in how computation is distributed. Dijkstra performs online optimization at every step, while RL shifts this cost to a one-time training phase ($\approx$ 40--68 minutes in our experiments). Consequently, RL policies provide a substantially lower deployment-time computational burden, making them more suitable for long-running or repeatedly deployed systems where inference cost dominates.

This observation also complements earlier observations where Dijkstra was shown to be consistently more cost-effective than the \textit{Threshold} heuristic in terms of total cost. While that result highlights the benefit of explicit planning over rule-based heuristics, the present analysis shows that this improvement comes at a significant computational cost. In practice, this introduces a trade-off between decision quality (planning) and runtime efficiency (learned policies), particularly in settings where frequent re-planning is required.

\subsection{Constraint mechanisms}
%Further analysis on constraint mechanisms for RL
%HERE

% -----------------------------------------------------------------------
% TABLE 2: CONSTRAINT MECHANISMS (Oracle + Estimated merged)
% -----------------------------------------------------------------------
\begin{table*}[h]
\centering
\caption{DQN and PPO performance with constraint mechanisms: reward shaping and safety mask, under oracle and estimated dynamics.}
\label{tab:constraints}
\scriptsize
\setlength{\tabcolsep}{4pt}
\resizebox{\textwidth}{!}{%
\begin{tabular}{l | ll | ll | ll | ll | ll}
\toprule
\multirow{2}{*}{Method} &
  \multicolumn{2}{c|}{$c_\text{f} = 100$} &
  \multicolumn{2}{c|}{$c_\text{f} = 500$} &
  \multicolumn{2}{c|}{$c_\text{f} = 1000$} &
  \multicolumn{2}{c|}{$c_\text{f} = 5000$} &
  \multicolumn{2}{c}{$c_\text{f} = 10000$} \\
 & Cost & Fail & Cost & Fail & Cost & Fail & Cost & Fail & Cost & Fail \\
\midrule
\multicolumn{11}{l}{\textit{Oracle dynamics}} \\
\midrule
DQN shaped & 8134 $\pm$ 0     & 6.0 $\pm$ 0.0 & 10534 $\pm$ 0    & 6.0 $\pm$ 0.0 & 13534 $\pm$ 0    & 6.0 $\pm$ 0.0 & 37534 $\pm$ 0        & 6.0 $\pm$ 0.0 & 65744 $\pm$ 1373  & 4.7 $\pm$ 1.2 \\
DQN masked & 33703 $\pm$ 0    & 5.0 $\pm$ 0.0 & 35703 $\pm$ 0    & 5.0 $\pm$ 0.0 & 40928 $\pm$ 3853 & 5.0 $\pm$ 0.0 & 58577 $\pm$ 529   & 4.7 $\pm$ 0.5 & 83458 $\pm$ 361   & 4.3 $\pm$ 0.9 \\
PPO shaped & 20710 $\pm$ 5377 & 5.1 $\pm$ 0.8 & 30087 $\pm$ 4023 & 4.1 $\pm$ 0.2 & 28033 $\pm$ 3620 & 3.9 $\pm$ 0.7 & 47048 $\pm$ 6364     & 2.2 $\pm$ 2.1 & 55429 $\pm$ 4298  & 1.5 $\pm$ 1.7 \\
PPO masked & 56810 $\pm$ 8411 & 2.0 $\pm$ 1.6 & 56979 $\pm$ 8237 & 2.2 $\pm$ 0.8 & 54924 $\pm$ 1549 & 2.3 $\pm$ 0.9 & 58846 $\pm$ 2265 & 0.0 $\pm$ 0.0 & 73921 $\pm$ 15265 & 1.1 $\pm$ 1.4 \\
\midrule
\multicolumn{11}{l}{\textit{Estimated dynamics}} \\
\midrule
DQN shaped & 8134 $\pm$ 0     & 6.0 $\pm$ 0.0 & 10534 $\pm$ 0    & 6.0 $\pm$ 0.0 & 13534 $\pm$ 0    & 6.0 $\pm$ 0.0 & 37534 $\pm$ 0        & 6.0 $\pm$ 0.0 & 67534 $\pm$ 0    & 6.0 $\pm$ 0.0 \\
DQN masked & 37317 $\pm$ 7734 & 4.8 $\pm$ 0.2 & 36135 $\pm$ 3102 & 4.9 $\pm$ 0.1 & 40570 $\pm$ 3418 & 5.0 $\pm$ 0.0 & 60956 $\pm$ 3883 & 4.1 $\pm$ 0.7 & 83913 $\pm$ 5322  & 3.5 $\pm$ 0.2 \\
PPO shaped & 17770 $\pm$ 2751 & 5.3 $\pm$ 0.3 & 21585 $\pm$ 3433 & 5.1 $\pm$ 0.3 & 17697 $\pm$ 3448 & 5.6 $\pm$ 0.5 & 39737 $\pm$ 858      & 5.3 $\pm$ 0.3 & 65265 $\pm$ 3810  & 4.5 $\pm$ 0.5 \\
PPO masked & 52595 $\pm$ 3628 & 1.7 $\pm$ 1.3 & 53469 $\pm$ 9318& 1.5 $\pm$ 1.2 & 51516 $\pm$ 4356 & 2.7 $\pm$ 1.8 & 62374 $\pm$ 1219 & 1.8 $\pm$ 0.9 & 65448 $\pm$ 8061  & 1.2 $\pm$ 1.0 \\
\bottomrule
\end{tabular}
}
\end{table*}

We further investigate the effect of constraint mechanisms for RL agents in Table~\ref{tab:constraints}, specifically focusing on reward shaping and action masking. The goal of this analysis is to understand whether explicit modifications to the training signal or action space can improve failure avoidance compared to vanilla, unconstrained RL strategies.

Across both DQN and PPO, reward shaping does not consistently alter the underlying failure behavior; DQN's performance remains largely unchanged while PPO shows moderate and inconsistent shifts in the cost–failure trade-off across settings. This suggests that scaling the failure penalty alone is insufficient to reliably enforce safety-oriented behavior in this environment.

In contrast, action masking leads to a more pronounced reduction in failures across both algorithms, particularly for PPO, where the mask can substantially reduce failure occurrences in some configurations. However, this improvement comes at the cost of significantly increased total cost, reflecting a more conservative, i.e., frequent, maintenance policy rather than an improved scheduling strategy.

Overall, the results indicate that action-space constraints are more effective than reward-based modifications in influencing failure behavior, but neither mechanism is sufficient to consistently match the zero-failure performance of planning and heuristic across all settings. This reinforces the observation that reliability constraints are difficult to enforce implicitly through standard RL formulations without explicit structural restrictions on decision-making.

\subsection{Further analysis of DQN failures}\label{subsec:dqn_analysis}
% -----------------------------------------------------------------------
% TABLE 3: DQN GAMMA ABLATION (Oracle + Estimated merged)
% -----------------------------------------------------------------------
\begin{table*}[h!]
\centering
\caption{Ablation study: Effect of discount factor on DQN with continuous and discrete observations under oracle dynamics.}
\label{tab:gamma_ablation}
\scriptsize
\setlength{\tabcolsep}{4pt}
\resizebox{\textwidth}{!}{%
\begin{tabular}{l | ll | ll | ll | ll | ll}
\toprule
\multirow{2}{*}{Method} &
  \multicolumn{2}{c|}{$c_\text{f} = 100$} &
  \multicolumn{2}{c|}{$c_\text{f} = 500$} &
  \multicolumn{2}{c|}{$c_\text{f} = 1000$} &
  \multicolumn{2}{c|}{$c_\text{f} = 5000$} &
  \multicolumn{2}{c}{$c_\text{f} = 10000$} \\
 & Cost & Fail & Cost & Fail & Cost & Fail & Cost & Fail & Cost & Fail \\
\midrule
\multicolumn{11}{l}{\textit{Oracle dynamics}} \\
\midrule
DQN cont. \ $\gamma=0.99$  & 8134 $\pm$ 0 & 6.0 $\pm$ 0.0 & 10534 $\pm$ 0 & 6.0 $\pm$ 0.0 & 13534 $\pm$ 0 & 6.0 $\pm$ 0.0 & 37534 $\pm$ 0 & 6.0 $\pm$ 0.0 & 66856 $\pm$ 959  & 5.7 $\pm$ 0.5 \\
DQN cont.\ $\gamma=0.999$ & 8134 $\pm$ 0 & 6.0 $\pm$ 0.0 & 10534 $\pm$ 0 & 6.0 $\pm$ 0.0 & 13534 $\pm$ 0 & 6.0 $\pm$ 0.0 & 37534 $\pm$ 0 & 6.0 $\pm$ 0.0 & 66795 $\pm$ 578  & 5.3 $\pm$ 0.5 \\
DQN disc.\ $\gamma=0.99$  & 8134 $\pm$ 0 & 6.0 $\pm$ 0.0 & 10534 $\pm$ 0 & 6.0 $\pm$ 0.0 & 13534 $\pm$ 0 & 6.0 $\pm$ 0.0 & 37534 $\pm$ 0 & 6.0 $\pm$ 0.0 & 65895 $\pm$ 1159 & 5.3 $\pm$ 0.5 \\
DQN disc.\ $\gamma=0.999$ & 8134 $\pm$ 0 & 6.0 $\pm$ 0.0 & 10534 $\pm$ 0 & 6.0 $\pm$ 0.0 & 13534 $\pm$ 0 & 6.0 $\pm$ 0.0 & 37534 $\pm$ 0 & 6.0 $\pm$ 0.0 & 67534 $\pm$ 0   & 6.0 $\pm$ 0.0 \\
\bottomrule
\end{tabular}
}
\end{table*}

Next, we conduct an ablation study on DQN to investigate why it achieves lower overall cost than PPO despite exhibiting higher failure rates and minimal variance across random seeds. In particular, we aim to understand why DQN appears largely insensitive to the value of $c_f$ and what type of policy it ultimately learns. To do so, we repeat the experiments using both continuous and discrete state representations, as well as a larger discount factor ($\gamma = 0.999$).

Inspection of the learned policies shows that, across almost all configurations in Table~\ref{tab:gamma_ablation}, DQN converges to a near zero-maintenance policy, leading to highly consistent behavior across seeds. In contrast, PPO does not exhibit the same degree of collapse and instead maintains more stochastic and comparatively failure-averse behavior. This pattern is observed for both continuous and discrete state representations. Furthermore, increasing $\gamma$ from 0.99 to 0.999 produces little qualitative change, with DQN continuing to converge to essentially the same policy in most settings.

We posit that this behavior arises from a discount--horizon mismatch in the problem formulation. In typical RL formulations that optimize for a discounted cumulative reward, future failure penalties are geometrically attenuated by $\gamma^t$, effectively yielding a planning horizon on the order of $\frac{1}{1-\gamma}$ steps. In our setting, bearing failures typically occur far beyond this effective horizon, causing the discounted value of preventing failure to become negligible relative to the immediate cost of maintenance. Consequently, the optimization objective structurally favors policies that avoid short-term maintenance costs even at the expense of long-term failures. We emphasize that this discount-horizon argument is a plausible heuristic explanation rather than a formal proof, though a similar observation has been independently documented in other sequential decision-making domains \cite{wireless2025average}. However, other factors, including reward scaling, exploration schedule, function approximation error, target-network update frequency, replay-buffer composition, and the relative magnitude of maintenance and failure costs, may also contribute to DQN's observed policy collapse. Formally disentangling these effects is left to future work.

Increasing $\gamma$ extends the effective horizon in principle, but does not fully resolve this mismatch in practice. Even at $\gamma = 0.999$, late-stage failures remain heavily discounted relative to maintenance costs, and the optimization landscape continues to favor non-intervention policies. As a result, DQN exhibits qualitatively similar behavior across discount factors and state representations, although some variance begins to emerge at the highest failure cost ($c_f = 10000$).

Taken together, these results suggest that DQN's superior cost performance does not stem from improved maintenance scheduling, but rather from convergence to a degenerate yet stable zero-maintenance policy. This also explains the negligible variance across seeds: once the policy reaches this attractor region, exploration no longer meaningfully alters the learned behavior.

\subsection{Discussion and broader impact}

Building on the preceding sections, we reflect on the broader implications of these results. This study is not intended to introduce a novel method, but to use a controlled bearing maintenance setting to expose the practical challenges of applying advanced decision-making approaches in PHM. A central objective is to highlight the complexity of choosing between RL and planning-based methods for predictive maintenance, where each approach operates under different assumptions and incurs different trade-offs in computation, reliability, and total operational cost.

A key takeaway is that planning and RL are effectively solving different problems. Planning enforces reliability as a hard constraint, while RL treats failures as a soft penalty. Under long horizons, discounting weakens the impact of delayed failures, which can bias RL toward short-term cost minimization rather than long-term reliability. As a result, lower cost in RL does not necessarily imply better decisions, but often reflects a fundamentally different objective formulation.

We emphasize that this asymmetry is a deliberate feature of our experimental design rather than an incidental limitation. By holding the environment, cost model, and evaluation protocol fixed while allowing the reliability requirement itself to differ across paradigms, we isolate the effect of objective formulation on the resulting policies. Consequently, Dijkstra's zero-failure performance should be read as a direct consequence of its imposed hard constraint rather than as an emergent empirical advantage of planning over RL in general. A matched-objective comparison, for instance a constrained RL formulation evaluated against a finite-penalty planner, would be needed to isolate any residual differences attributable to the algorithms themselves rather than to their objectives; we view this as an important direction for future work.

Several limitations should be noted. The study is conducted on a fixed dataset and a fixed-horizon setting, using a relatively small fleet of six bearings drawn from a single run-to-failure dataset, and the observed behaviors may change under different data distributions, system scales, or operating horizons. Both planning and RL agents are also assumed to observe accurate remaining-useful-life estimates or discrete health states derived from ground-truth failure times. Real deployments operate on uncertain RUL estimates produced by a prognostics model, and performance under such estimation noise remains to be characterized. In addition, RL results are sensitive to training dynamics, evaluation protocols, and reward design, and reported performance may reflect locally optimal policies instead of globally optimal solutions. RL results are also averaged over only three random seeds without formal statistical significance testing, and some reported differences, particularly for PPO under reward shaping and action masking, fall within one standard deviation. We also did not explore more advanced constrained RL formulations, which could enforce reliability requirements more directly, such as CMDPs, average-reward RL, or risk-sensitive RL. Furthermore, we did not exhaust the range of planning approaches, as our study considers only a Dijkstra-based planning method. More sophisticated planning algorithms may offer different cost-reliability trade-offs or improved computational efficiency. Evaluating alternative planning methods alongside more advanced RL formulations remains an important direction for future work.

\section{Conclusion}

This study highlights distinct formulation differences between planning and RL for multi-asset maintenance: planning enforces reliability as a hard constraint, while vanilla RL optimizes a discounted expected cost that may tolerate failures. DQN consistently collapses to a zero-maintenance policy due to a mismatch between its discount horizon and the longer asset lifetimes, making failure prevention negligible relative to immediate maintenance costs. PPO partially mitigates this issue but only achieves near-zero failure behavior with additional constraint mechanisms. Notably, both heuristic and planning policies achieve zero failures, though planning achieves a much lower operational cost, at the expense of significantly higher computational time. Overall, the results suggest that planning and RL are complementary: planning suits strict reliability requirements, while RL is better when limited failures are acceptable and long-term efficiency matters. Future work should explore constrained MDPs, scaling behavior, and hybrid planning–RL approaches.

\bibliographystyle{unsrtnat}
\bibliography{references}  
\end{document}